\documentclass[letterpaper]{article} 
\usepackage{aaai24}  
\usepackage{times}  
\usepackage{helvet}  
\usepackage{courier}  
\usepackage[hyphens]{url}  
\usepackage{graphicx} 
\usepackage{natbib}  
\usepackage{caption} 
\def\eg{\emph{e.g.}} 

\def\ie{\emph{i.e.}}

\def\vs{\emph{vs. }}
\def\al{{et al. }}

\usepackage{algorithm}
\usepackage{algorithmic}

\usepackage{newfloat}
\usepackage{listings}

\usepackage{array}
\usepackage{booktabs}
\usepackage{multirow}
\usepackage{graphicx}                                             
\usepackage{float} 
\usepackage{amssymb}
\usepackage{amsmath}
\usepackage{dsfont}
\usepackage[table]{xcolor}
\usepackage{color}

\usepackage{newfloat}
\usepackage{listings}
\DeclareCaptionStyle{ruled}{labelfont=normalfont,labelsep=colon,strut=off} 
\floatstyle{ruled}
\newfloat{listing}{tb}{lst}{}
\floatname{listing}{Listing}
\nocopyright

\title{Adaptive Integration of Partial Label Learning and Negative Learning for Enhanced Noisy Label Learning}
\author{
    Mengmeng Sheng\textsuperscript{\rm 1},
    Zeren Sun\textsuperscript{\rm 1}\thanks{Corresponding Author},
    Zhenhuang Cai\textsuperscript{\rm 1},
    Tao Chen\textsuperscript{\rm 1},
    Yichao Zhou\textsuperscript{\rm 1},
    Yazhou Yao\textsuperscript{\rm 1\:\!\textasteriskcentered}
}
\affiliations{
    \textsuperscript{\rm 1}Nanjing University of Science and Technology, Nanjing, China\\
    \{shengmengmemg, zerens, czh, taochen, yczhou, yazhou.yao\}@njust.edu.cn

}

\usepackage{bibentry}

\begin{document}

\maketitle

\begin{abstract} 
There has been significant attention devoted to the effectiveness of various domains, such as semi-supervised learning, contrastive learning, and meta-learning, in enhancing the performance of methods for noisy label learning (NLL) tasks.
However, most existing methods still depend on prior assumptions regarding clean samples amidst different sources of noise (\eg, a pre-defined drop rate or a small subset of clean samples).
In this paper, we propose a simple yet powerful idea called \textbf{NPN}, which revolutionizes \textbf{N}oisy label learning by integrating \textbf{P}artial label learning (PLL) and \textbf{N}egative learning (NL).
Toward this goal, we initially decompose the given label space adaptively into the candidate and complementary labels, thereby establishing the conditions for PLL and NL.
We propose two adaptive data-driven paradigms of label disambiguation for PLL: hard disambiguation and soft disambiguation.
Furthermore, we generate reliable complementary labels using all non-candidate labels for NL to enhance model robustness through indirect supervision.
To maintain label reliability during the later stage of model training, we introduce a consistency regularization term that encourages agreement between the outputs of multiple augmentations.
Experiments conducted on both synthetically corrupted and real-world noisy datasets demonstrate the superiority of NPN compared to other state-of-the-art (SOTA) methods.
The source code has been made available at {\color{purple}{\url{https://github.com/NUST-Machine-Intelligence-Laboratory/NPN}}}.

\end{abstract}

\section{Introduction}
\begin{figure}[t]
\centering
\includegraphics[width=\linewidth]{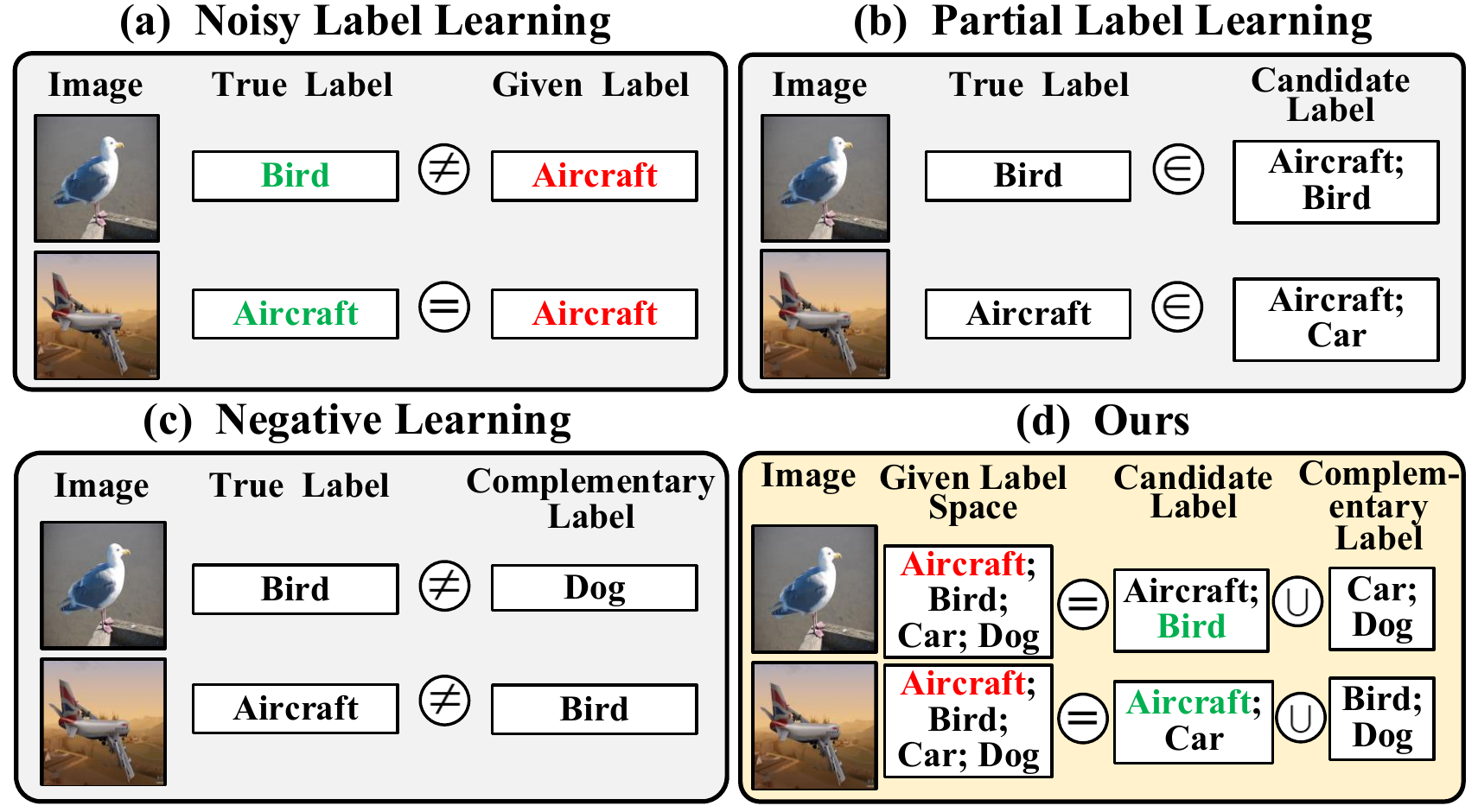}
\vspace{-0.6cm}
\caption{(a-c) The difference between Noisy Label Learning, Partial Label Learning, and Negative Learning. (d) Our decomposition of the given label space to candidate label
and complementary label to establish the necessary conditions for Partial Label Learning and Negative Learning.}
\label{figure_motivation}
\vspace{-0.5cm}
\end{figure}

Over the past decades, the revolution in supervised learning has been driven by large-scale, well-annotated datasets such as ImageNet \cite{ImageNet}. 
However, collecting these datasets has become a bottleneck due to its expense and time-consuming nature, hindering the scalability of models. 
The acquisition of large-scale datasets furnished with annotations of high quality for supervised learning algorithms is a formidable challenge.
To address this issue, weakly supervised learning \cite{Weakly_supervised_learning, Chen_1, Chen_2} has gained considerable attention, which includes, but is not limited to, noisy label learning \cite{dividemix,co-teaching,sun2022pnp}, multi-label learning \cite{multi-label_learning1,multi-label_learning2,multi-label_learning3}, partial label learning \cite{partial_label_learning1,partial_label_learning2}, and semi-supervised learning \cite{chen2023softmatch, FreeMatch}.
In this paper, our primary focus is on one specific weakly supervised learning problem known as noisy label learning.
NLL involves handling low-quality samples with noisy labels (as shown in Fig.~\ref{figure_motivation} (a)), which mainly stem from crowd-sourcing platforms \cite{Crowd-sourcing} or web image search engines \cite{Sesrch_engine} employed in dataset construction.
The powerful learning capability of deep neural networks allows them to fit any noisy labels, leading to inferior performance in image classification tasks. 
Consequently, developing robust methods to mitigate the impact of noisy labels assumes paramount importance.

The existing NLL methods primarily focus on combating label noise by incorporating strategies such as semi-supervised learning, contrastive learning, and meta-learning~\cite{Liu_2,Liu_1}.
Semi-supervised learning (SSL) often constitutes an appropriate choice to leverage the selected noisy samples through sample selection rather than abandoning them. 
For example, DivideMix \cite{dividemix} employs a mixture model to characterize the per-sample loss distribution, dynamically partitioning the training data into a labeled subset comprising clean samples and an unlabeled subset comprising noisy samples. 
The model is subsequently trained on both labeled and unlabeled data, adopting a semi-supervised approach.
UNICON \cite{UNICON} and NCE \cite{NCE} propose refined sample selection strategies while incorporating semi-supervised learning.
However, these methods tend to neglect the reliability of the pseudo-labels generated by the semi-supervised approach, which may lead to correction bias.

Another line of the NLL methods boosts performance by integrating contrastive learning.
This approach facilitates feature extraction that remains uninfluenced by label noise, thereby demonstrating promising results.
Jo-SRC \cite{josrc} takes predictions from two distinct views of each sample to estimate its ``likelihood'' of being clean or out-of-distribution in a contrastive learning manner.
Sel-CL \cite{Sel-CL} proposes selective-supervised contrastive learning combined with a joint loss to enhance model generalization performance by introducing consistency regularization.
Nevertheless, these methods are sensitive to hyper-parameters. 
Recently, Meta-learning has also been employed to mitigate label noise.
L2RW \cite{L2RW} proposes a meta-learning instance re-weighting approach to assign instance weights to noisy instances.
MLC \cite{MLC} proposes to adopt a label correction network as a meta-model to produce corrected labels for noisy samples.
However, these meta-learning methods tend to require prior knowledge (\eg, a small subset of clean samples).

To tackle the above challenges, we propose an innovative approach named \textbf{NPN} for noisy label learning.
To be specific, NPN synergistically incorporates two distinct learning paradigms: partial label learning and negative learning, to combat label noise effectively.
PLL enables each training example to be labeled with a coarse candidate set, which is well-suit for real-world data annotation scenarios with label ambiguity (as shown in Fig.~\ref{figure_motivation} (b)). 
NL provides indirect supervision information using a complementary label that teaches the network ``the input image does not belong to this complementary label'' (as shown in Fig.~\ref{figure_motivation} (c)). 
Thus, as a preliminary step, we decompose the given label space to the candidate and complementary labels to establish the conditions for PLL and NL (as shown in Fig.~\ref{figure_motivation} (d)). 
For PLL, we propose two paradigms of label disambiguation in an adaptive data-driver manner: hard disambiguation and soft disambiguation.
Additionally, we suggest generating reliable complementary labels using all non-candidate labels for NL to enhance model robustness through indirect supervision.
Finally, a consistency regularization term is applied to improve both feature extraction and model prediction.
We provide comprehensive experimental results and extensive ablation studies to validate the effectiveness and superiority of NPN on synthetically corrupted and real-world datasets.
Our contributions can be summarized as follows:

(1) We present a simple yet powerful method, named NPN, to alleviate negative impacts induced by noisy labels. NPN introduces a paradigm shift in conventional noisy label learning by amalgamating partial label learning and negative learning. This involves partitioning the given label space into candidate labels for PLL and complementary labels for NL.

(2) We introduce two paradigms of label disambiguation for PLL: hard disambiguation (NPN-hard) and soft disambiguation (NPN-soft). To enhance the effectiveness of NL, we advocate using all reliable non-candidate labels as the complementary label, as opposed to randomly selecting a single unreliable non-given label.

(3) Comprehensive evaluations on one synthetically corrupted and three real-world noisy datasets show that NPN outperforms SOTA methods in addressing noisy labels. Extensive ablation studies are conducted to further verify the effectiveness of the proposed method.

\section{Related Work}

\textbf{Noisy Label Learning.}
Recently, numerous methods have been proposed for learning with noisy labels \cite{josrc, LiuTongLiang_v2_TPAMI, HANBO_v1_TPAMI}.
They focus on combating noisy labels through sample selection, label correction, or noise regularization.
Previous literature about sample selection attempts to detect noisy labels by exacerbating the natural resistance of neural networks to noise.
BARE \cite{BARE} proposes an adaptive sample selection strategy that relies only on batch statistics to provide robustness against label noise.
Another line of research focuses on label correction, which typically attempts to rectify sample labels using the model predictions.
Jo-SRC \cite{josrc} uses the temporally averaged model (\ie, mean-teacher model) to generate reliable pseudo-label distributions for training.
Besides, PLS \cite{PLS} proposes considering the confidence of corrected labels.
Certain studies in the literature emphasize focus on noise regularization, such as mixup \cite{mixup}, a dedicated loss term \cite{ELR}, or contrastive learning.
Unsupervised regularization \cite{Co-LDL, Representation} has also been shown effective in improving the classification accuracy of neural networks trained on noisy datasets.
\begin{figure*}[t]
\centering
\includegraphics[width=\linewidth]{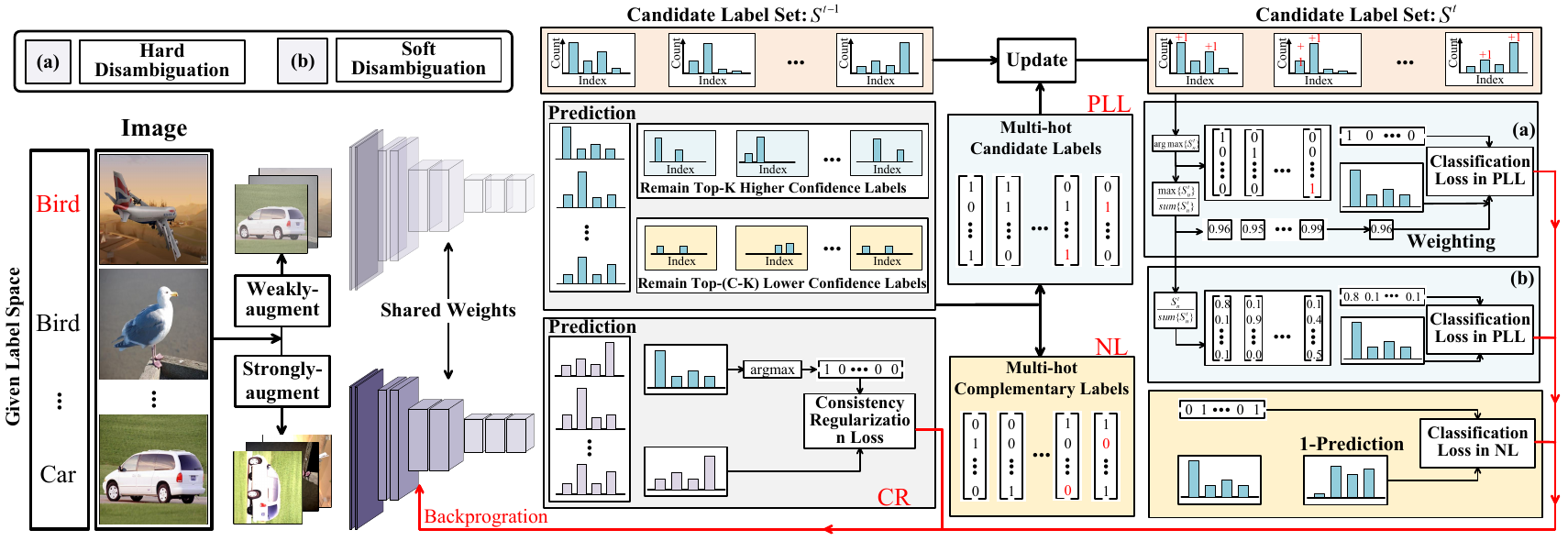}
\vspace{-0.5cm}
\caption{The overall framework of NPN. We first decompose the given label space into the candidate and complementary labels based on model predictions. For PLL, we introduce two adaptive, data-driven paradigms of label disambiguation: hard disambiguation and soft disambiguation. To enable adaptive, data-driven disambiguation, we maintain a candidate label set $\mathcal{S}$ that stores the candidate labels for each sample during the training process. For NL, we incorporate reliable complementary labels, which encompass all non-candidate labels, to enhance the model's robustness through indirect supervision. Finally, NPN introduces a consistency regularization term that promotes prediction agreement between sample augmentations.}
\label{figure1}
\vspace{-0.3cm}
\end{figure*}
\\
\textbf{Partial Label Learning.}
Partial Label Learning allows each training example to be annotated with a candidate label set, in which the ground-truth label is guaranteed to be included.
Two typical strategies have been proposed for label disambiguation in PLL: the average-based and identification-based disambiguation strategies.
The average-based disambiguation strategy treats each candidate label equally during model learning \cite{PLL}. 
The identification-based disambiguation strategy considers the ground-truth label as a latent variable. 
It identifies the ground-truth label by deriving confidence scores for all candidate labels \cite{PLL_1, PLL2}.
These label disambiguation strategies can serve as inspiration for noisy label learning, which can be regarded as a variation of PLL where candidate labels encompass all categories in the dataset.
\\
\textbf{Negative Learning.}
Negative Learning is an indirect learning method where CNNs are trained by utilizing a complementary label that indicates ``input image does not belong to this complementary label.''
NLNL \cite{NLNL} introduces a noise-robust learning approach that utilizes negative learning to mitigate the risk of learning incorrect information.
JNPL \cite{JNPL} improves the method in \cite{NLNL} and proposes a novel approach that combines both negative and positive learning.
UPS \cite{UPS} employs negative learning on noisy and low-confidence labels to generalize negative pseudo-labels, which can be utilized for multi-label classification as well as negative learning to improve single-label classification. 
In this work, we propose to exploit the indirect supervision information by negative learning.

\section{Methods}

\subsection{Problem Statement}
Throughout this paper, we mainly focus on noisy label learning for image classification. 
Considering the problem of a $\mathcal{C}$-class classification task, we denote the noisy training set $\mathcal{D}_{train}=\{(x_n,y_n) | n=1,...,N\}$ as the input with $N$ training samples.
$x_n$ denotes the $n$-th image, and $y_n \in \{0,1\}^C$ denotes the corresponding one-hot label of $C$ classes.
The label $y_n$ may not be equivalent to the ground-truth label denoted by $y_n^*$, which is not observable during training.
Suppose a classification neural network $\mathcal{F}(\cdot,\theta)$ maps the input space to the $\mathcal{C}$-dimensional score space $\mathcal{F: X \to \mathbb{R}^C}$, where $\theta$  denotes network parameters.
We further denote $p^c(x_n, \theta)$ as the predicted softmax probability of training sample $x_n$ over its $c$-th class.
The goal is to train $\mathcal{F}(\cdot,\theta)$ on the noisy training set $\mathcal{D}_{train}$ to perform accurate prediction on $\mathcal{D}_{test}$, where $\mathcal{D}_{test}=\{(x_m, y_m^*)|m=1,...,M \}$ is the test set with accurate labels. 
The loss function in the conventional training scheme is the cross-entropy loss as follows:
\begin{equation}\label{eq_1}
\mathcal{L}_{ce} = -\frac{1}{N}\sum_{n=1}^{N}\sum_{c=1}^{C}y_n^clog(p^c(x_n, \theta)).
\end{equation}

The existing literature \cite{LiuTongLiang_v6_ICML, HANBO_v3_ICLR, LiuTongLiang_v5_CVPR} highlights that optimizing networks using the cross-entropy loss on noisy datasets often results in inferior performance.
In this work, we aim to enhance traditional noisy label learning by integrating partial label learning and negative learning methodologies.
PLL allows each training example $x_n$ to be labeled with the coarse candidate label set $\mathcal{Y}_n$.
NL is an indirect training method for CNNs, where the non-given complementary label $\tilde{\mathcal{Y}}_n$ is employed. 
It is worth noting that $y_n$ is encoded as a single label using one-hot encoding, whereas $\mathcal{Y}_n$ and $\tilde{\mathcal{Y}}_n$ are encoded using multi-hot encoding in our NPN.

\subsection{Label Space Decomposition}
In the previous literature of NLL, more efforts have been devoted to semi-supervised learning, contrastive learning, and meta-learning. 
However, methods following semi-supervised learning tend to overlook the reliability of the pseudo-labels generated by the semi-supervised approach, which may lead to correction bias \cite{dividemix}.
Methods employing contrastive learning methods are considerably sensitive to hyper-parameters \cite{josrc}. 
Meta-learning methods struggle in requiring prior knowledge (\ie, a small subset of clean data) \cite{L2RW}.

To mitigate the above issues, we propose to revolutionize conventional noisy label learning by integrating partial label learning and negative learning.
As an initial step, we decompose the given label space into a combination of the candidate label set in PLL and the complementary label in NL.
This can prevent the model from overfitting noisy labels through the direct and indirect supervision information learned from PLL and NL, respectively.
We construct the candidate label set in PLL by selecting the given label $y_n$ and the category with the highest prediction confidence $\hat{y}_n \in \{0,1\}^C$ as follows:
\begin{equation}\label{eq_2}
\mathcal{Y}_n =  y_n + \hat{y}_n,\  \hat{y}_n^k=\mathds{1}_{k=\mathop{\arg \max}\limits_{c\in \{1,..,C\}}p^{c}(x_n, \theta)}.
\end{equation}
$\mathds{1}_A$ is an indicator function, which equals 1 if $A$ is true, and 0 otherwise.

We further generate complementary labels for NL using the remaining non-candidate labels in the given label space.
Unlike NLNL \cite{NLNL} and JNPL \cite{JNPL}, which randomly select a single label from non-given labels as the complementary label, we use all non-candidate labels as the complementary label set,
\begin{equation}\label{eq_3}
\tilde{\mathcal{Y}}_n =  \neg{\mathcal{Y}_n} = \mathcal{I} - \mathcal{Y}_n,
\end{equation}
where $\mathcal{I}$ denotes the entire label space.

\noindent
\textbf{Discussion}.
Notably, the candidate labels within PLL should encompass the ground-truth label, while the complementary label within NL is the exact opposite.
Consequently, upholding the reliability of both candidate and supplementary labels becomes paramount.
We compare the probability of true labels among the top-k predicted categories on CIFAR100N, as shown in Fig.\ref{figure2}.
Fig.\ref{figure2} (a) shows that the true label frequently falls between the given label and the label obtained the highest prediction confidence.
In this case, our NPN achieves the highest test accuracy, as shown in Fig.\ref{figure2} (b).
The candidate labels with higher confidence further reinforce the reliability of complementary labels used for NL, owing to the relationship $\tilde{\mathcal{Y}}_n +\mathcal{Y}_n = \mathcal{I}$.

\noindent
\subsection{Label Disambiguation in Partial Label Learning}
The main challenge of PLL arises from the fact that the ground-truth label is concealed within the candidate label set and is not directly accessible to the learning algorithms.
Thus, the primary focus of PLL is to tackle the task of candidate label disambiguation.
Existing methods predominantly concentrate on two strategies: average-based disambiguation and identification-based disambiguation.
The average-based disambiguation strategy treats each candidate label equally during the training phase.
It computes the average of all the modeling outputs associated with each candidate label during the testing phase \cite{PLL}.
On the other hand, the identification-based disambiguation strategy treats the ground-truth label as a latent variable \cite{PLL_1}. 
It aims to identify the true label by assigning different confidence scores to the candidate labels.

In this work, we propose two adaptive, data-driven paradigms of label disambiguation for PLL: hard disambiguation (\ie, identification-based disambiguation) and soft disambiguation (\ie, average-based disambiguation).
Specifically, we first design a candidate label set $\mathcal{S}$ to store the candidate labels for each sample during the training process as:
\begin{equation}\label{eq_4}
\mathcal{S}_{n}^t=\left\{
                        \begin{array}{ll}
                            y_n, &  t=0\\
                            \mathcal{S}_{n}^{t-1}+\mathcal{Y}_n^t, & t>0 
                        \end{array}.
\right.
\end{equation}
We denote $\mathcal{S}_{n}^t$ as the frequency distribution of the candidate labels of the $n$-th sample among $C$ classes, which has been observed in the previous $t$ epochs.
$\mathcal{Y}_n^t$ represents the aforementioned multi-label candidate label of the $n$-th sample in the $t$-th epochs.

The sample with the highest frequency of occurrence in the candidate label set $\mathcal{S}_{n}^t$ is more likely to be the ground-truth sample, allowing for the disambiguation of candidate labels.
Following the idea of hard disambiguation, we construct $\tilde{y}$ as the disambiguation labels and compute $\mathcal{L}_{PLL}$ as follows:
\begin{equation}\label{eq_5}
\tilde{y}_n    = \mathop{\arg \max}\limits_{j = 1,..,C} \mathcal{S}_{n}^t,
\end{equation}
and
\begin{equation}\label{eq_6}
\mathcal{L}_{PLL}^{hard} = -\frac{1}{N}\sum_{n=1}^{N}\frac{max\{ \mathcal{S}_{n}^t\}}{sum\{ \mathcal{S}_{n}^t\}}\tilde{y}_nlog(p(x_n, \theta)).
\end{equation}
We utilize $\frac{max\{ \mathcal{S}_{n}^t\}}{sum\{ \mathcal{S}_{n}^t\}}$ as the weight to measure the reliability of the disambiguated labels, aiming to mitigate the impact of inevitable errors from partially disambiguated samples.

Furthermore, we propose a soft disambiguation strategy:
\begin{equation}\label{eq_7}
\tilde{y}_n    =  \frac{\mathcal{S}_{n}^t}{sum\{\mathcal{S}_{n}^t\}},
\end{equation}
and
\begin{equation}\label{eq_8}
\mathcal{L}_{PLL}^{soft} = -\frac{1}{N}\sum_{n=1}^{N}\sum_{c=1}^{C}\tilde{y}_n^clog(p^c(x_n, \theta)).
\end{equation}
The core of the soft label disambiguation strategy revolves around individually considering each label within the candidate label set and constructing soft labels based on their respective frequencies, thereby accomplishing effective label disambiguation.

\noindent
\textbf{Discussion}.
The effectiveness of PLL is directly influenced by the purity of the labels achieved through disambiguation.
To further validate the impact of our label disambiguation strategies, we present the overall precision of label disambiguation \vs epochs in Fig.\ref{figure2} (c). 
The results demonstrate that selecting more labels into candidate labels increases the difficulty of label disambiguation, subsequently resulting in a decline in test accuracy. 
This rationale elucidates our decision to choose the given label and category with the highest prediction confidence as candidate labels $\mathcal{Y}$ during each epoch. 
Thus, we ensure that the candidate labels encompass the ground-truth labels to the fullest extent possible while concurrently alleviating the complexities associated with label disambiguation.

\subsection{Indirect Supervision in Negative Learning}
Negative learning is generally carried out to assist the model in acquiring indirect information by randomly selecting a non-given label as the complementary label, as stated in NLNL and JNPL.  
Moreover, the complementary label should differ from the ground-truth label.
However, the ground-truth label is present among the non-given labels when the given label is noisy.
They encounter the challenge of potentially mis-selecting the ground-truth label as the complementary label.
This becomes particularly pronounced in high-noise scenarios.

As mentioned above, the labels more likely to be ground-truth labels are often assigned to the candidate labels.
The remaining labels, less likely to be the ground-truth labels, are thus assigned to the complementary labels.
In addition, the complementary labels $\tilde{\mathcal{Y}}$ that we propose encompass a set of labels, as opposed to a randomly selected non-given label.
It offers the model a richer source of supervision information, concurrently mitigating the impact of occasional mis-selection of true labels.
Finally, we calculate the negative learning loss using the constructed complementary labels as follows:
\begin{equation}\label{eq_9}
\mathcal{L}_{NL} = -\frac{1}{N}\sum_{n=1}^{N}\sum_{c=1}^{C}\tilde{\mathcal{Y}}_n^clog(1-p^c(x_n, \theta)).
\end{equation}
Eq.\eqref{eq_9} facilitates the optimization of the probability value associated with the complementary label to approach zero. This adjustment leads to an increase in the probability values of other classes, aligning with the objectives of NL.

\begin{figure*}[t]
\vspace{-0.3cm}
	\raggedright
	\includegraphics[width=0.98\linewidth]{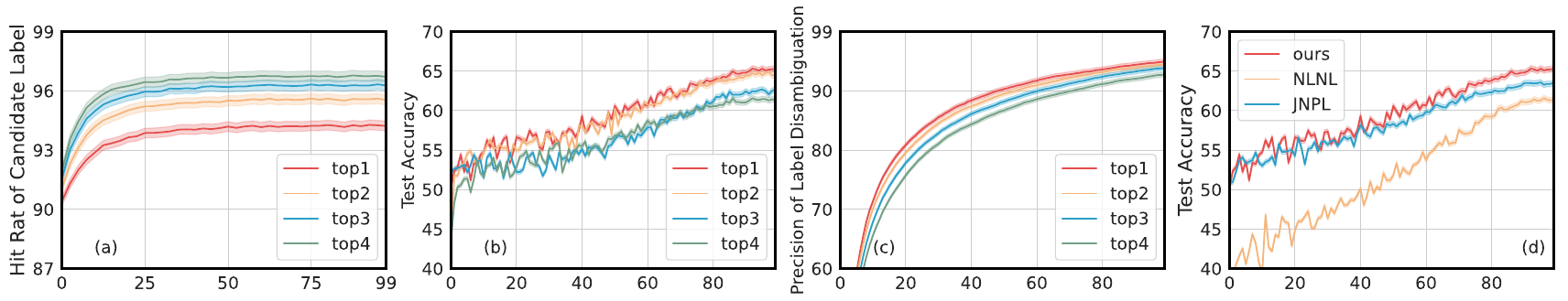}
 \vspace{-0.15cm}
	\caption{Comparison of label disambiguation using different top-k predicted labels (a-c) and comparison of different negative learning methods (d). (a) The overall hit rate of ground-truth labels within candidate labels (\%) \vs epochs. (b) The test accuracy (\%) \vs epochs. (c) The overall precision of label disambiguation (\%) \vs epochs. (d) The test accuracy (\%) of using different negative learning methods (\%) \vs epochs.}
	\label{figure2}
\vspace{-0.4cm}
\end{figure*}

\noindent
\textbf{Discussion}.
To address the two drawbacks of NLNL and JNPL, we primarily focus on two aspects:
(1) Our NL framework aims to discover more reliable (\ie, lower prediction probability) complementary labels.
(2) By utilizing all non-candidate labels as the complementary label set rather than a single one, we can boost the performance of NL.
As shown in Fig.\ref{figure2} (d), NPN attains superior performance when compared to previous approaches (\ie, NLNL and JNPL).
\begin{algorithm}[t]
\footnotesize{
\caption{Our proposed NPN algorithm}
\begin{flushleft}
\textbf{Input:} The training set $D_{train}$, network $\theta$, warm-up epochs $E_{w}$, total epochs $E_{total}$, batch size $bs$.
\end{flushleft}
\begin{algorithmic}[1] 
 \FOR {$epoch=1,2,\ldots,E_{total}$}
 \IF {$epoch \le E_{w}$}
    \FOR {$iteration=1,2,\ldots$}
    \STATE Fetch $B=\{(x_i, y_i)\}^{bs}$ from $D_{train}$
    \STATE Calculate $\mathcal{L}_{ce} = - \sum_{i=1}^{bs}y_i\ \mathcal{\log}\ p(x_i,\theta)$ 
    \STATE Update $\theta$ by optimizing $\mathcal{L}_{ce}$
    \STATE Obtain candidate label $\mathcal{Y}$ using Eq.~\eqref{eq_2}
    \STATE Update candidate label set $\mathcal{S}$ using Eq.~\eqref{eq_4}
    \ENDFOR
    \ENDIF
 \IF {$E_{w} < epoch \le E_{total}$}
    \FOR {$iteration=1,2,\ldots$}
    \STATE Obtain candidate label $\mathcal{Y}$ using Eq.~\eqref{eq_2}
    \STATE Obtain complementary label $\tilde{\mathcal{Y}}$ using Eq.~\eqref{eq_3}
    \STATE Update candidate label set $\mathcal{S}$ using Eq.~\eqref{eq_4}
    \STATE Calculate $\mathcal{L}_{PLL}$ using Eq.~\eqref{eq_6} or Eq.~\eqref{eq_7}
    \STATE Calculate $\mathcal{L}_{NL}$ using Eq.~\eqref{eq_9} 
    \STATE Calculate $\mathcal{L}_{REG}$ using Eq.~\eqref{eq_11} 
    \STATE Calculate $\mathcal{L} = \mathcal{L}_{PLL} + \alpha \mathcal{L}_{NL} + \beta\mathcal{L}_{REG}$
    \STATE Update $\theta$ by optimizing $\mathcal{L}$
    \ENDFOR
    \ENDIF
\ENDFOR
\end{algorithmic}
\begin{flushleft}
\textbf{Output:} Updated network $\theta$.
\end{flushleft}
\label{alg}  }
\end{algorithm}
\noindent
\subsection{Overall Framework}
In summary, we merge partial label learning and negative learning to cope with noisy samples without demanding any strong prior assumptions.
This approach is implemented in an adaptive, data-driven manner, capitalizing on both the direct supervision in PLL and the indirect supervision in NL to fortify the robustness of the model.
The overall learning procedure of our NPN is illustrated in Fig.~\ref{figure1} and Algorithm \ref{alg}.
The objective loss function in our NPN is as follows:
\begin{equation}\label{eq_10}
\mathcal{L} = \mathcal{L}_{PLL} + \alpha \mathcal{L}_{NL} + \beta \mathcal{L}_{REG},
\end{equation}
where $\alpha$ and $\beta$ are weighting factors. 

$\mathcal{L}_{REG}$ denotes the consistency regularization (CR) loss, which encourages prediction consistency between weakly- ($A_w$) and strongly-augmented ($A_s$) views of samples.
This loss term implicitly enhances the robustness of the model by regularizing it as follows:
\begin{equation}\label{eq_11}
\mathcal{L}_{REG}= -\frac{1}{N}\sum_{n=1}^{N}\hat{y}_nlog(p(A_s(x_n), \theta)),
\end{equation}
in which $\hat{y}_n=\mathop{\arg \max}\limits_{j = 1,..,C} p^j(A_w(x_n), \theta)$.

\begin{table*}[t]
\centering
\setlength{\tabcolsep}{3.8mm}{
\begin{tabular}{cccccccccccc}
\toprule
\multirow{2}{*}{\textbf{Methods}} & 
\multicolumn{4}{c}{\textbf{Symmetric}} & \multicolumn{4}{c}{\textbf{Asymmetric}} \\  \cmidrule(r){2-5} \cmidrule(r){6-9}
  & 10\% & 20\% & 40\% & 80\% & 10\% & 20\% & 40\% & 50\% \\ 
\hline
Standard &  43.27  & 35.50 & 21.02 & 3.84 & 45.89 & 40.86 & 28.43 & 23.40\\
Decoupling (Malach \al 2017)  & 50.39  & 43.84 & 33.08 & 9.33 & 51.59 & 46.36 & 33.58 & 27.30 \\
Co-teaching \cite{co-teaching} & 57.69 & 56.21 & 52.09 & 22.83  & 57.31 & 52.78 & 37.26 & 25.99 \\
Co-teaching+ \cite{Co-teaching+} & 54.99  & 52.87 & 45.64 & 18.55 & 54.26 & 51.25 & 38.78 & 27.46\\
JoCoR \cite{JoCoR}  & 60.75  & 58.69 & 52.16 & 14.18 & 59.38 & 56.10 & 38.24 & 22.41 \\
Co-LDL \cite{Co-LDL} & 61.55  &59.73  & 53.56  & 25.12 & 61.05 & 60.40 & 52.28 & 30.50\\
SPRL $^{\dagger}$ \cite{SPRL} & 56.11  & 53.64  & 48.84  & 22.30 & 55.74 & 56.07 & 49.48 & 26.61\\
AGCE $^{\dagger}$ \cite{AGCE} & 60.25  & 59.38  & 54.47  & 27.41 & 60.34 & 59.07 & 43.04 & 24.78\\
\hline
\rowcolor{gray!20} \textbf{NPN-soft}  & 64.75 & 62.76 & 58.80 & 31.69 & 64.52 & 63.55 & 57.11 & \textbf{33.23}\\
\rowcolor{gray!20} \textbf{NPN-hard} & \textbf{66.79} & \textbf{65.27} &	\textbf{61.35} & \textbf{36.88} & \textbf{67.19} & \textbf{66.36} & \textbf{60.11} & 33.19\\
\bottomrule
\end{tabular}}
\vspace{-0.2cm}
\caption{Average test accuracy (\%) on CIFAR100N over the last ten epochs. Experiments are conducted under various noise conditions (“Symmetric” and “Asymmetric” denote the symmetric and asymmetric label noise, respectively).}
\label{tab_1}
\end{table*}

\section{Experiments}
In this section, we evaluate our NPN by comparing it with SOTA methods. We also conduct ablation studies to validate the effectiveness of each design choice in NPN.
The experiments are conducted on a synthetically corrupted dataset, namely CIFAR100N \cite{CIFAR}, as well as three real-world datasets, namely Web-Aircraft, Web-Car, and Web-Bird \cite{sun2021webly}.
Experiments follow the setup employed in \cite{Co-LDL}.

\noindent
\subsection{Experiment Setup}
\textbf{Synthetically Corrupted Dataset.}
We conduct experiments on CIFAR100N, which is derived from the CIFAR100 dataset.
CIFAR100 consists of 50,000 training images and 10,000 test images, divided into 100 clean-annotated classes.
We consider two approaches to generating noisy labels:
(1) Symmetric noise: Labels are randomly corrupted from their ground-truth classes to other ones with a probability of $\frac{n}{C-1}$, where $n$ means the noise rate.
(2) Asymmetric noise: Labels are corrupted from ground-truth classes to the next classes with the probability $n$.
\\
\textbf{Real-World Datasets.}
We also conduct experiments on real-world datasets (\ie,  Web-Aircraft, Web-Car, and Web-Bird). 
These datasets consist of training images obtained through web image search engines, resulting in inevitable label noise.
Compared to the synthetically corrupted dataset, real-world datasets exhibit a more complex and realistic challenge in practical scenarios. 
They contain various types of label noise (\ie, symmetric noise, asymmetric noise, and open-set noise).
\\
\textbf{Experiment Details.}
Following \cite{Co-LDL}, we use a seven-layer CNN network as the backbone for CIFAR100N.
The network is trained using an SGD optimizer with a momentum of 0.9 for 300 epochs.
During the first 100 epochs, a fixed learning rate (\ie, 0.005) is used for warm-up training, followed by 200 epochs of training using a cosine-decay learning rate (starting from 0.005) for robust training.
The batch size is 128.
For real-world datasets, we leverage ResNet50 \cite{Resnet} pre-trained on ImageNet as our backbone.
The network is also trained using an SGD optimizer with a momentum of 0.9 for 100 epochs.
The batch size is 32, and the learning rate is 0.001.
The learning rate decays in a cosine annealing manner. 
\\
\textbf{Baselines}
We compare NPN with the following SOTA methods for CIFAR100N: Decoupling \cite{Decoupling}, Co-teaching \cite{co-teaching}, Co-teaching+ \cite{Co-teaching+}, JoCoR, Co-LDL, SPRL \cite{SPRL} and AGCE \cite{AGCE}.
Besides, we compare NPN with the following SOTA methods for real-world datasets: PENCIL \cite{PENCIL}, DivideMix \cite{dividemix} AFM \cite{AFM}, and Self-adaptive \cite{Self-adaptive}).
$^{\dagger}$ means that we re-implement the method using its open-sourced code and default hyper-parameters.
Additionally, we provide results of using noisy training directly for training (denoted as Standard).

\begin{table*}[t]
\centering
\setlength{\tabcolsep}{5mm}{
\begin{tabular}{ccccccc}
\toprule
\multirow{2}{*}{\textbf{Methods}}  & \multirow{2}{*}{\textbf{Backbone}}  & 
\multicolumn{4}{c}{\textbf{Performances(\%)}} \\  \cline{3-6} 
& & Web-Aircraft 	& Web-Bird    & Web-Car  & Average \\ 
\hline
Standard & ResNet50 & 60.80  & 64.40 & 60.60 &61.93\\
Decoupling (Malach \al 2017) & ResNet50 & 75.91 & 71.61 & 79.41 & 75.64\\
Co-teaching \cite{co-teaching}  & ResNet50 & 79.54 & 76.68 & 84.95 & 80.39\\
Co-teaching+ \cite{Co-teaching+}  & ResNet50 & 74.80 & 70.12 & 76.77 & 73.90\\
PENCIL \cite{PENCIL}  & ResNet50 & 78.82 & 75.09 & 81.68 & 78.53\\
JoCoR \cite{JoCoR}  & ResNet50 & 80.11 & 79.19 & 85.10 & 81.47\\
DivideMix (Li \al 2020)  & ResNet50 & 82.48 & 74.40 & 84.27 & 80.38 \\
AFM \cite{AFM}  & ResNet50 & 81.04 & 76.35 & 83.48 & 80.29\\
Self-adaptive (Huang \al 2020) & ResNet50 & 77.92 & 78.49 & 78.19 & 78.20\\
Co-LDL \cite{Co-LDL} & ResNet50 & 81.97 & 80.11 & 86.95 & 83.01\\
SPRL $^{\dagger}$ \cite{SPRL}   & ResNet50 & 80.77 & 80.41  & 86.17  & 82.45 \\
AGCE $^{\dagger}$ \cite{AGCE}  & ResNet50 & 84.22  & 75.60  & 85.16  & 81.66\\
\hline
\rowcolor{gray!20}\textbf{NPN-soft}  & ResNet50 & 83.65 & 79.36 & 85.46 & 82.82\\ 
\rowcolor{gray!20}\textbf{NPN-hard}  & ResNet50 & \textbf{86.02} & \textbf{80.91} & \textbf{88.26} &\textbf{85.06} \\
\bottomrule
\end{tabular}}
\vspace{-0.2cm}
\caption{Comparison with SOTA approaches in test accuracy (\%) on real-world noisy datasets: Web-Aircraft, Web-Bird, Web-Car. Results of existing methods are mainly drawn from \cite{Co-LDL}. }
\label{tab_3}
\vspace{-0.3cm}
\end{table*}

\subsection{Evaluation on Synthetically Corrupted Datasets}
The performance comparison on CIFAR100N is summarized in Table~\ref{tab_1}, clearly demonstrating the consistent superiority of both NPN-hard and NPN-soft approaches.
Results of existing methods shown in Table~\ref{tab_1} are mainly obtained from Co-LDL \cite{Co-LDL}.
Our proposed NPN-hard outperforms all the competing methods across all noise types and noise rates. 
Similarly, NPN-soft also achieves a performance boost over existing methods, although it yields lower performance than NPN-hard in most cases.
In the most challenging scenario (\ie, Symmetric-80\%), NPN-hard still outperforms other methods, achieving an accuracy of 36.88\% (an 11.76\% improvement).
These experiments on CIFAR100N demonstrate the robustness of NPN in handling both symmetric and asymmetric label noise across various noise rates.

\subsection{Evaluation on Real-World Datasets}
In addition to the experimental results on synthetic datasets, Table~\ref{tab_3} compares NPN with SOTA methods on three real-world datasets.
Results of SOTA methods shown in Table~\ref{tab_3} are also drawn from Co-LDL \cite{Co-LDL}.
As shown in Table~\ref{tab_3}, our NPN-hard achieves accuracies of 86.02\%, 80.91\%, and 88.26\% on Web-Aircraft, Web-Bird, and Web-Car, respectively, surpassing Co-LDL \cite{Co-LDL} by 4.05\%, 0.80\%, and 1.31\%.
The results demonstrate the efficacy of our NPN in more challenging real-world scenarios.
It showcases its adaptive, data-driven nature that operates effectively without additional prior knowledge, such as a predefined drop rate or threshold.

\begin{table}[t]
\renewcommand\tabcolsep{10pt}
	\begin{center}
		\centering
		\begin{tabular}{cccccc}
			\toprule
			\textbf{Standard}       & \textbf{NL}  & \textbf{PLL}   & \textbf{CR}            	& \textbf{Test Accuracy} \\
			\midrule
			\checkmark			    & 		 		& 			    & 			  				& 35.50 \\
			\checkmark				& \checkmark	& 				& 			  				& 54.70 \\
			\checkmark				& \checkmark	& \checkmark	& 			  				& 60.62 \\
			\checkmark				& \checkmark	& \checkmark	& \checkmark			  	& 65.27 \\
			\bottomrule
		\end{tabular}
	\end{center}
\vspace{-0.4cm}
\caption{Effects of different ingredients in test accuracy (\%) on CIFAR100N (noise type and noise rate are ``symmetric'' and 20\%, respectively).}
\vspace{-0.6cm}
\label{tab_4}
\end{table}

\subsection{Ablation Studies}
\begin{figure}[t]
 \vspace{-0.1cm}
	\raggedright
	\includegraphics[width=\linewidth]{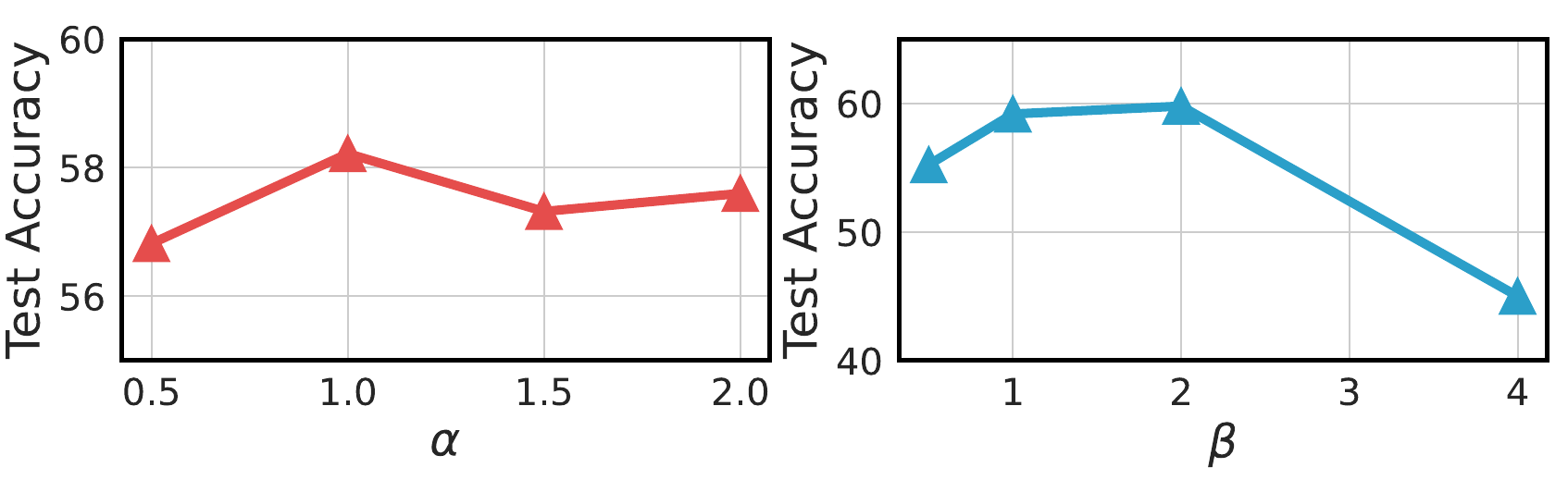}
 \vspace{-0.6cm}
	\caption{The hyper-parameter sensitivities of $\alpha$ and $\beta$.}
	\label{figure3}
\vspace{-0.4cm}
\end{figure}

In this subsection, we conduct experiments to investigate the effectiveness of each component in NPN.
Unless otherwise stated, ablation experiments are performed based on our NPN-hard on CIFAR100N (Sym-20\%).
Table~\ref{tab_4} and Fig.~\ref{figure3} show the results of the ablation experiments and provide insights into the contribution of each component in NPN. 
\\
\textbf{Effects of Partial Label Learning.}
As stated above, we focus on combating label noise by transforming noisy label learning into partial label learning and negative learning.
For PLL, we design a candidate label set $\mathcal{S}$ to store the candidate labels for each sample during the training process.
We evaluate two paradigms of label disambiguation based on the candidate label set: hard disambiguation and soft disambiguation.
The results in Table~\ref{tab_1} demonstrate that, in most cases, hard disambiguation outperforms soft disambiguation, except for a few extreme scenarios (\ie, Asymmetric-50\%). 
Furthermore, as shown in Table~\ref{tab_4}, incorporating PLL leads to a 5.92\% performance gain. 

To further validate the effect of label disambiguation, we present the comparison of label disambiguation among the top-k predicted labels.
Fig.\ref{figure2} (c) demonstrates that selecting more labels into candidate labels increases the difficulty of label disambiguation.
Consequently, it results in a decline in test accuracy, as shown in Fig.\ref{figure2} (a) and Fig.\ref{figure2} (b). 
This rationale elucidates our design to choose the given label and category with the highest prediction confidence as candidate labels during each epoch. 
\\
\textbf{Effects of Negative Learning.}
We generate the complementary label for NL and boost the model robustness by the indirect supervision of NL.
Different from the existing methods, we focus on mining more reliable complementary labels and utilizing all complementary labels to enhance the performance of NL.
Table~\ref{tab_4} illustrates that adopting NL boosts model performance by 19.2\% compared to the baseline method (\ie, Standard).
We also conduct experiments by replacing our complementary label generation with typical generation methods that randomly select a non-given label  (\ie, NLNL and JNPL) in Fig.\ref{figure2} (d).
It is evident that our NPN achieves the highest test accuracy.
\\
\textbf{Effects of Consistency Regularization.}
During the later stage of model training, it is inevitable that the model will overfit some noisy samples, which results in the inability to select reliable candidate labels and complementary labels.
Accordingly, we propose to impose additional consistency regularization to achieve enhancement in both feature extraction and model prediction. 
As shown in Table~\ref{tab_4}, the results demonstrate that consistency regularization successfully boosts the model performance by 4.65\%.
\\
\textbf{Effects of Hyper-parameters.}
Our NPN combines partial label learning and negative learning to effectively learn with noisy labels without strong prior assumptions.
In our approach, we only use hyper-parameters (\ie, $\alpha$ and $\beta$) to control the weights of different loss terms in Eq.~\eqref{eq_10}.
Fig.~\ref{figure3} provides an ablation study on the impact of different $\alpha$ and $\beta$ settings.
When $\alpha$ is 1.0, and $\beta$ is 2.0, NPN achieves the highest performance.

\section{Conclusion}
This paper introduced a simple yet effective framework, termed NPN, designed to learn with noisy labels. 
It integrated two other distinct learning paradigms, namely partial label learning and negative learning, to effectively combat label noise.
Different from traditional noisy label learning methods, NPN was implemented in an adaptive and data-driven manner.
We started by decomposing the given label space into a combination of candidate labels for PLL and complementary labels for NL.
For PLL, we proposed two adaptive, data-driver paradigms of label disambiguation: hard disambiguation and soft disambiguation.
Additionally, we suggested generating reliable complementary labels using all non-candidate labels for NL to enhance model robustness through indirect supervision.
Finally, we introduced a consistency regularization term that encouraged prediction agreement between different sample augmentations to improve both feature extraction and model prediction. 
Extensive experiments and ablation analysis confirmed the effectiveness and superiority of our proposed method.

\bibliography{aaai24}

\end{document}